\documentclass{article}
\usepackage[final]{corl_2026}
\usepackage{amsmath,amssymb,booktabs,graphicx}
\usepackage{fancyhdr}
\fancypagestyle{accepted}{%
  \fancyhf{}%
  \fancyhead[C]{\small\itshape Accepted at the CoRL 2026 Workshop on the Science of Physical AI Safety (SPAIS).}%
}
\graphicspath{{figures/}}
\makeatletter
\renewcommand{\@noticestring}{CoRL 2026 Workshop on the Science of Physical AI Safety (SPAIS).}
\makeatother
\title{When Black-Box Robot Policies Fail: Online Sensor Diagnosis and Repair with a Transparent Surrogate}
\author{Marc Ziegler$^{1}$ \quad Batu \"Ozen$^{1}$ \quad Sami Mentes$^{1}$ \quad Thomas Waschulzik$^{2}$ \quad Alois Knoll$^{1}$ \\
\normalfont $^{1}$Technical University of Munich \quad $^{2}$Siemens Mobility \\
\normalfont\texttt{marc.eric.ziegler@tum.de}}
\hypersetup{pdftitle={When Black-Box Robot Policies Fail: Online Sensor Diagnosis and Repair with a Transparent Surrogate},pdfauthor={Marc Ziegler, Batu \"Ozen, Sami Mentes, Thomas Waschulzik, Alois Knoll}}
\begin{document}
\maketitle
\thispagestyle{accepted}

\begin{abstract}
Robot foundation models may generalize across tasks and embodiments, but their actions still depend on sensors that can fail. When the policy is frozen or inaccessible, can an external monitor diagnose and repair corrupted inputs without access to its internals? We test a structured surrogate that separates periodic gait from sensor-dependent corrections to guide input repair around a frozen locomotion policy. On simulated Go2, offline diagnosis raises walking success from 259/640 to 429/640 episodes; online repair reaches 382/640, with no triggers in 64 held-out fault-free episodes. A matched boosted-tree surrogate predicts clean actions similarly well but yields no detected repair improvement over no repair. Separate fits improve walking on Go1 and ANYmal C, although phase-aware gradients perform better. Evidence remains limited to one policy and a partly screened simulated fault suite; robot foundation model and hardware validation remain open.
\end{abstract}

\section{Introduction}
Robot foundation models promise transfer across tasks and
embodiments~\citep{openx2024rtx,kim2024openvla}, yet their actions depend on
reliable sensor observations. When internals are inaccessible,
monitors see only inputs and actions. We study the frozen multi-embodiment
URMA locomotion policy~\citep{bohlinger2024onepolicy}. On Go2 at \(0.5\,\mathrm{m/s}\), a \(0.3\,\mathrm{rad/s}\) yaw-gyroscope
bias prevents walking in 16/16 episodes. Classical diagnosis uses plant or
sensor models~\citep{isermann2005fdi}, whereas gradient attribution and VLA
steering require policy access%
~\citep{shrikumar2017deeplift,sundararajan2017ig,
haon2025mechanistic,buurmeijer2026observing}. Runtime guardrails constrain
plans~\citep{ravichandran2026guardrails}, while masked training builds sensor
robustness into the policy~\citep{peri2024masked}. Unusual sensor readings may nevertheless result from legitimate disturbances.

Prior work distills locomotion into boosted trees or symbolic
expressions~\citep{acero2024locomotion}. Structured controllers and sparse
mixtures also expose control structure%
~\citep{srisuchinnawong2025gollum,vincze2025smose}. Our surrogate separates
periodic gait from sensor-dependent corrections, ranks abnormal input
contributions, and guides repair. The frozen policy produces every action,
so the surrogate never replaces the controller.

We contribute diagnosis and repair without policy gradients, paired offline
and online evaluations, and tests on three quadrupeds. This controlled setting motivates external supervision for robot foundation
models~\citep{stocking2026physicalai}. Direct validation on those models
remains open.

\section{Method}
\begin{figure}[t]
\centering
\includegraphics[width=0.90\linewidth,height=0.27\textheight,keepaspectratio]{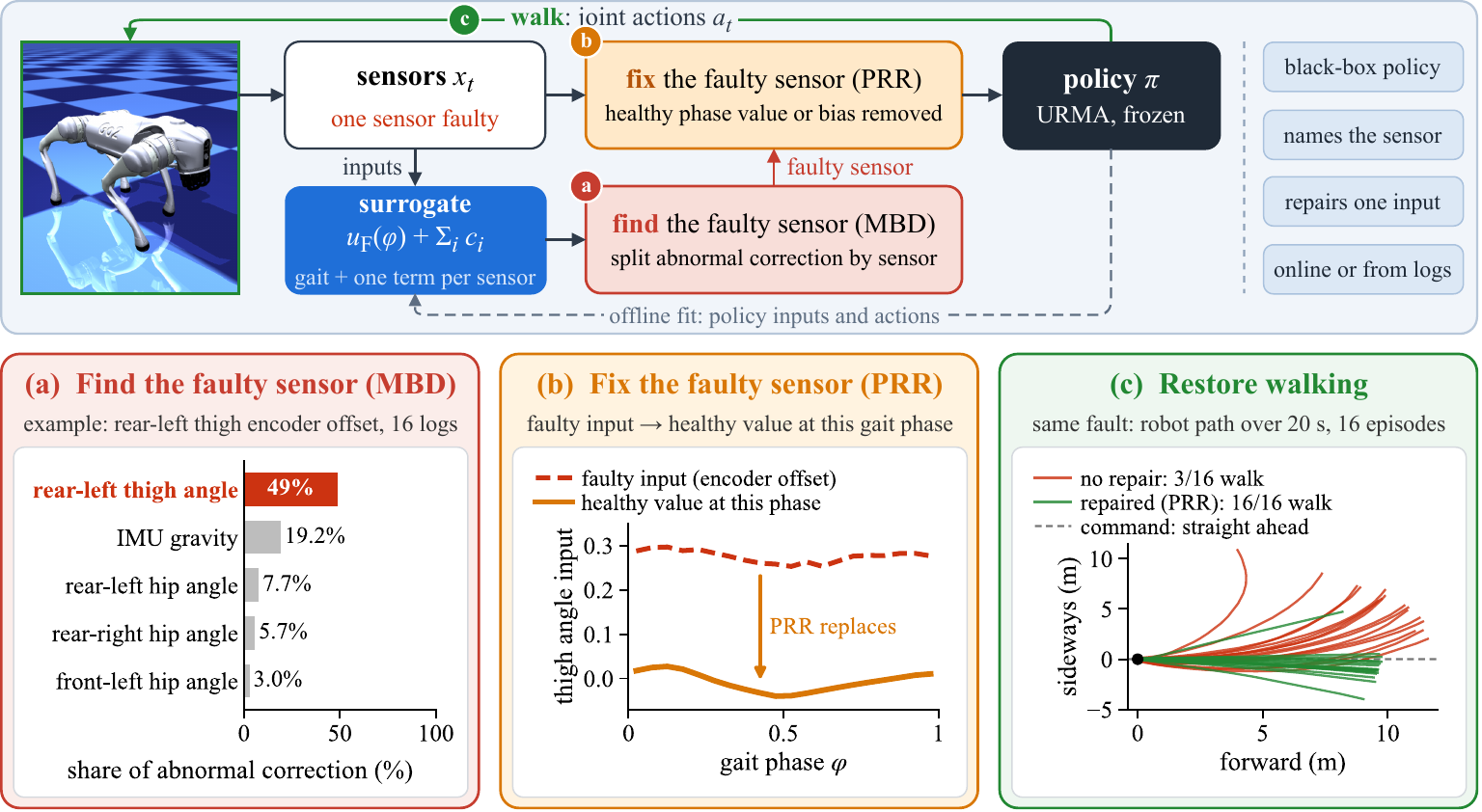}
\caption{\textbf{Method overview: Diagnosis and input repair.} MBD identifies the faulty sensor from the structured surrogate's per-sensor contributions, and PRR repairs its input before the frozen policy acts. For a \(+0.3\) rear-left thigh-encoder offset: (a) the sensor explains
49\% of the abnormal correction across 16 diagnosis logs; (b) PRR supplies
its phase-matched healthy value; and (c) repair removes the sideways drift
and raises walking success from 3/16 to 16/16 fresh episodes.}
\label{fig:overview}
\end{figure}

\paragraph{Structured action surrogate.}
Used only for diagnosis, the surrogate predicts 12 joint-action targets
\(u_t\) at command speed \(v\) (Figure~\ref{fig:overview}):
\begin{equation}
u_t =
u_{\rm F}(\varphi_t,v)
+\operatorname{clip}_{[-0.15,0.15]}(r_t),
\qquad
r_{t,j} =
b_j(\varphi_t,v)+\sum_i c_{i,t,j}.
\label{eq:surrogate}
\end{equation}
We represent the gait \(u_F\) using four Fourier harmonics. Both the gait and correction models are
conditioned on the commanded speed \(v\). The correction \(r_t\) combines a
phase-dependent baseline \(b_j(\varphi_t,v)\) with contributions from
individual input features. The contribution $c_{i,t,j}$ of standardized input
\(z_{i,t}\) to joint \(j\) is
\(c_{i,t,j}=z_{i,t}[a_{ij}(v)+a^{\mathrm{s}}_{ij}(v)\sin(2\pi\varphi_t)+a^{\mathrm{c}}_{ij}(v)\cos(2\pi\varphi_t)]\).
The 57 inputs describe base motion, gravity, joint states, previous actions,
contacts, command, and timing. The baseline and contributions exactly
reconstruct the surrogate's correction before clipping.

We collect 64,000 policy state--action pairs at four speeds, fit the gait,
then fit the remaining action error by ridge regression. Feedback fitting
filters startup and invalid-phase rows and holds out episodes. We also
collect 32,000 policy action targets at states reached under alternating
policy/surrogate control~\citep{ross2011dagger} to broaden feedback training.

Front-left touchdown resets the gait phase. Between touchdowns, the clock
advances using its latest period estimate; touchdown intervals of 8--60
steps update this estimate with averaging rate 0.2. Offline fitting and
fault-free reference construction use simulator-derived contacts. During
diagnosis and repair, phase is computed from the potentially corrupted
contact readings visible to the policy.

\paragraph{Diagnosis from abnormal corrections.}
Mechanism-based diagnosis (MBD) uses the surrogate's additive decomposition
to identify which sensor contributions drive abnormal corrections. We
discretize gait phase into 20 equal bins (5\% of a cycle each); this
resolution was fixed before fault evaluation. From fault-free trajectories,
we estimate in each bin \(\beta\) the mean and standard deviation of each
joint correction, \(\bar r_j(\beta)\) and \(s_j(\beta)\), and the mean
contribution of each input, \(\bar c_{i,j}(\beta)\). At time \(t\), joint
\(j\) is flagged when its correction deviates from the phase-matched
reference by more than two standard deviations:
\(\mathcal F=\{(t,j): |r_{t,j}-\bar r_j(\beta_t)|>2s_j(\beta_t)\}\).
For each sensor unit \(U\), MBD then computes
\begin{equation}
S_U
=
\sum_{(t,j)\in\mathcal F}\sum_{i\in U}
\bigl(c_{i,t,j}-\bar c_{i,j}(\beta_t)\bigr)
\operatorname{sgn}\bigl(r_{t,j}-\bar r_j(\beta_t)\bigr).
\label{eq:mechanism}
\end{equation}
Thus, contributions that push a flagged correction farther in its abnormal
direction increase the sensor's score, whereas opposing contributions
decrease it. MBD ranks 34 sensor units: the gyroscope, gravity vector,
12 joint positions, 12 joint velocities, four contacts, and four contact
ages.

Offline diagnosis aggregates scores across 16 faulty trajectories, excluding
the first 25 steps and the final 50 steps before a fall. Although evaluation
assumes that a fault is present, MBD abstains when no corrections are flagged.

\paragraph{Baselines.}
\emph{Phase deviation} measures departure from each sensor's healthy value
at the same phase: $|x_i-\bar x_i(\beta)|/s_i^x(\beta)$. It averages over
channels in each sensor unit and sums over time. A gated variant uses only
steps where MBD flags an abnormal correction.
\emph{Phase-aware gradients} require policy internals. They compute
$(x_i-\bar x_i(\beta))^2\|\partial\pi/\partial x_i\|_2^2/12$
per channel, subtract its healthy phase-conditioned baseline, average over
time, and sum across channels in each sensor unit.
\emph{GBM--TreeSHAP} uses twelve boosted regressors~\citep{friedman2001gbm}.
Both fits use the same 12,740 clean and 7,827 alternating-control rows,
with identical inputs, splits, normalization, nominal gait, residual target,
and correction bound.
TreeSHAP uses tree-path-dependent
expectations~\citep{lundberg2020tree}. We group its attributions by sensor
and apply Eq.~\eqref{eq:mechanism} with its own anomaly mask. All methods
share sensor groups, diagnosis seeds, repair episodes, and PRR.

\paragraph{Input repair and online detection.}
Phase-referenced repair (PRR) changes only the selected unit $U$. For a
sensor that varies with phase, it supplies the healthy standardized mean,
$\hat z_{U,t}=\bar z_U(\beta_t)$. Otherwise, it estimates and subtracts bias:
\begin{equation}
\hat z_{U,t}=z_{U,t}-\hat b_{U,t},\qquad
\hat b_{U,t}
=0.99\hat b_{U,t-1}
+0.01\left[z_{U,t}-\bar z_U(\beta_t)\right],
\quad \hat b_{U,0}=0.
\label{eq:repair}
\end{equation}
The score $\kappa_U=|U|^{-1}\sum_{i\in U}\operatorname{SD}_{\beta}[\bar z_i(\beta)]$ measures how strongly a sensor normally varies over the gait cycle. If $\kappa_U\geq0.4$, PRR replaces its reading with the healthy reference for the current phase. Otherwise, it estimates and subtracts an additive bias. We selected the threshold using development data. Contacts and contact ages are replaced with constant nominal values.
Repaired values return to policy input units before the next policy call.
All repair rules were fixed before evaluation.

Offline MBD diagnoses prior faulty trajectories; repair runs in fresh
episodes. Online MBD starts after 25 steps and measures the fraction of
flagged joint--time pairs in nonoverlapping 50-step windows at
$50\,\mathrm{Hz}$. From the fifth window onward, it triggers when the lower
quartile over all completed windows exceeds a calibrated threshold.
We calibrate on 128 fault-free episodes in total. Each noise regime's
threshold is the maximum of this statistic over its calibration episodes
and decision times. Evaluation supplies the true noisy/noiseless regime.
At the first trigger, accumulated scores select a sensor; repair starts
on the next step and lasts until episode end. Runtime needs no policy
weights, gradients, or additional policy calls.

\section{Experiments}
\paragraph{Protocol.}
URMA controls Go2 in MuJoCo for 20 seconds at $50\,\mathrm{Hz}$ from
randomized starts, commanded at $v=0.5\,\mathrm{m/s}$. Pushed episodes
receive a $250\,\mathrm{N}$ force in a random horizontal direction for
$0.2$ seconds at six seconds. Survival means completing the episode.
Walking also requires mean velocity in each planar axis, after the first
two seconds, to be within $\max(0.05\,\mathrm{m/s},0.25v)$ of its command.
The 37 injected faults comprise 22 main cases and 15 extensions: offsets,
stuck readings, drift, dropout, spikes, and contact errors. During method development, we selected 10 main cases based on preliminary MBD performance. We report this development-informed subset separately from the remaining cases. Diagnosis uses 16 trajectories per fault (seeds 2100--2115).
Repair uses fresh seeds 2120--2135 on 20 main faults excluding the phase
anchor, with/without push: 640 paired episodes per method.
Fitting and calibration data are separate from diagnosis and repair data.
Diagnosis and repair share healthy references. Contacts are simulated
policy inputs, and the fault suite is a constructed stress test, not a
sample of hardware failure frequencies. Only online tests assess detection.
We report two-sided exact McNemar tests, unadjusted for multiple comparisons,
and a bootstrap over the 20 repaired fault types; inference is conditional
on this suite.

\paragraph{Diagnosis and repair.}
MBD identifies 27/37 faults: 8/10 screened main, 9/12 remaining main, and
10/15 extensions. Phase-aware gradients identify 33/37 ($p=0.11$ versus
MBD); phase deviation identifies 8/37 ($p=2.1\times10^{-5}$ versus MBD).
MBD-guided repair walks in 429/640 episodes versus 408/640 for phase-aware
gradients (38 MBD-only, 17 gradient-only; $p=0.0065$). True sensor selection
reaches 451/640 (Table~\ref{tab:results}). Against phase deviation, paired
walking favors MBD by 202 to 21 ($p=2.5\times10^{-38}$); gated and ungated
deviation give identical repair outcomes.
Screened faults show larger gains (92/288 to 219/288 over nine repaired
types) than the other eleven (167/352 to 210/352). The overall gain is
26.6 percentage points (exploratory 95\% fault-cluster bootstrap interval
11.7--41.7; 20,000 resamples), conditional on suite selection.

Despite similar clean-action $R^2$ (0.9764 versus 0.9766), matched
GBM--TreeSHAP identifies 5/37 faults and walks in 277/640 episodes, not
significantly above no repair ($p=0.076$) and below MBD
($p=1.2\times10^{-33}$). Reused faults and differing masks limit
attribution-specific conclusions.
\begin{table}[t]
\centering\small
\setlength{\tabcolsep}{4pt}
\caption{Results across three embodiments. For Go2, \(R^2\) measures action
prediction on 16 clean trajectories, and Diagnosed reports top-1 sensor
identification across 37 faults. Go2 Walk and Survive are counts over 640
episodes. Online MBD detects, diagnoses, and repairs faults within each
episode. Go1 and ANYmal C report walking over 352 episodes per method. On
these two robots, the GBM predicts actions directly and shares neither
MBD's gait model nor its input representation.}
\label{tab:results}
\begin{tabular}{lcccccc}
\toprule
& \multicolumn{4}{c}{Go2} & \multicolumn{2}{c}{Walk} \\
\cmidrule(lr){2-5}\cmidrule(lr){6-7}
Method & $R^2$ & Diagnosed & Walk & Survive & Go1 & ANYmal C \\
\midrule
No repair & -- & -- & 259 & 373 & 197 & 162 \\
MBD + PRR & 0.9766 & 27 & \textbf{429} & 467 & 305 & 265 \\
Online MBD + PRR & -- & -- & 382 & 419 & -- & -- \\
GBM--TreeSHAP + PRR & 0.9764 & 5 & 277 & 361 & 266 & 239 \\
Phase-aware gradient + PRR & -- & \textbf{33} & 408 & 455 & 319 & 283 \\
Phase deviation + PRR & -- & 8 & 248 & 361 & -- & -- \\
Ground-truth sensor + PRR & -- & 37 & 451 & 494 & -- & -- \\
\bottomrule
\end{tabular}
\end{table}
\paragraph{Online operation and anchor-contact faults.}
Online MBD + PRR raises walking from 259/640 to 382/640 (128 gains,
5 losses; $p=6.1\times10^{-32}$). It triggers in 454/640 faulty episodes,
at median step 274 ($5.48\,\mathrm{s}$ from episode start, among triggers),
and in 0/64 held-out fault-free episodes (Wilson 95\% upper bound 5.7\%).
Faults are injected from the first step, and step 274 is the earliest
permitted decision; 444 of the 454 triggers occur there.
The fault-free set includes noise and pushes: 35/64 walk, including all
16 clean, unpushed episodes.
A deviation detector counts readings beyond two healthy phase-conditioned
standard deviations. Its threshold is the maximum one-second outlier
fraction over the same calibration episodes. It never triggers and leaves
walking at 259/640; other thresholds and detectors remain untested.
For three front-left anchor-contact faults, PRR walks in 71/96 episodes
versus 76/96 without repair ($p=0.18$); healthy counterparts walk in 75/96.
These faults do not reduce aggregate walking on this block, and repair
adds no benefit.

\paragraph{Other embodiments.}
Separate fits on Go1 and ANYmal C test 13 shared faults at
$0.5\,\mathrm{m/s}$. Their gait and
phase-modulated feedback are fitted jointly, without Go2's clipping or
alternating-control data. Separate healthy screens select push magnitudes.
Repair tests 11 non-anchor faults with/without push:
352 episodes per robot. MBD diagnoses 6/13 and 8/13 faults, respectively,
and improves walking on both (Table~\ref{tab:results}). Phase-aware
gradients reach 602/704 versus MBD's 570/704 (42 gradient-only, 10 MBD-only;
$p=9.1\times10^{-6}$). Go2's method ranking does not transfer.
\section{Discussion and Conclusion}
MBD is the central diagnostic contribution, while PRR provides a controlled
way to test whether its diagnoses support effective intervention. The
structured decomposition makes the surrogate's sensor contributions
traceable without claiming to reveal the frozen policy's internal mechanism. A matched boosted-tree surrogate achieves similar
clean-action accuracy but does not provide comparable diagnostic or repair
performance. Clean-action accuracy alone is therefore insufficient evidence
of diagnostic value.
Our evidence is limited to one policy in simulation, a partly screened fault
suite, separately fitted surrogates for each robot, and simulator-derived
references. Future work should evaluate independently specified faults,
cross-robot transfer, robust phase estimation, abstention, simultaneous
faults, and onboard latency. Extension to manipulation and
vision--language--action policies also remains open.

\clearpage
{\footnotesize\bibliography{go2-corl}}
\end{document}